\pdfoutput=1
\documentclass{article}
\usepackage[preprint]{neurips_2021}
\usepackage[utf8]{inputenc} 
\usepackage[T1]{fontenc}    
\usepackage{hyperref}       
\usepackage{url}            
\usepackage{booktabs}       
\usepackage{amsfonts}       
\usepackage{nicefrac}       
\usepackage{microtype}      
\usepackage{xcolor}         

\usepackage{bm}
\usepackage{epsfig}
\usepackage{graphicx}
\usepackage{amsmath}
\usepackage{amssymb}
\usepackage{enumitem}
\usepackage{pifont}
\usepackage{multirow}
\usepackage{diagbox}

\usepackage{wrapfig}

\title{VidFace: A Full-Transformer Solver for Video Face Hallucination with Unaligned Tiny Snapshots}

\author{
	Yuan Gan\textsuperscript{1}\footnotemark[1]
	\quad
	Yawei Luo\textsuperscript{1}\footnotemark[1]
	\quad
	Xin Yu\textsuperscript{2}
	\quad
	Bang Zhang\textsuperscript{3}
	\quad
	Yi Yang\textsuperscript{1}
	\\ \\
	{$^1$Zhejiang University} \\ \small{$^2$University of Technology Sydney}\\ \small{$^3$DAMO Academy, Alibaba Group}
}

\begin{document}

\maketitle

\begin{abstract}
    In this paper, we investigate the task of hallucinating an authentic high-resolution (HR) human face from multiple low-resolution (LR) video snapshots. We propose a pure transformer-based model, dubbed VidFace, to fully exploit the full-range spatio-temporal information and facial structure cues among multiple thumbnails.
    Specifically, VidFace handles multiple snapshots all at once and harnesses the spatial and temporal information integrally to explore face alignments across all the frames, thus avoiding accumulating alignment errors. 
    Moreover, we design a recurrent position embedding module to equip our transformer with facial priors, which not only effectively regularises the alignment mechanism but also supplants notorious pre-training.
    Finally, we curate a new large-scale video face hallucination dataset from the public \emph{Voxceleb2}~\cite{chung2018voxceleb2} benchmark, which challenges prior arts on tackling unaligned and tiny face snapshots. To the best of our knowledge, we are the first attempt to develop a unified transformer-based solver tailored for video-based face hallucination. Extensive experiments on public video face benchmarks show that the proposed method significantly outperforms the state of the arts.
    
\end{abstract}

	\renewcommand{\thefootnote}{\fnsymbol{footnote}}
	\footnotetext[1]{Equal contribution.}
	\renewcommand*{\thefootnote}{\arabic{footnote}}
	
\section{Introduction}
Video-based face hallucination (VFH) targets at recovering a high-resolution (HR) human face from low-resolution (LR) video frames. This task has gained long-standing interest in the research community~\cite{liu2007face,wang2014comprehensive,liu2021point,yu2016ultra,zhang2021recursive,liu2014feature} due to its wide applications such as emotion recognition~\cite{hassan2019human,liu2014facial} and video surveillance~\cite{zhou2019anomalynet}. As a domain-specific task in video super-resolution, VFH is far less explored due to its high complexity and requirement in jointly modeling both spatio-temporal and facial prior information. 


Recent studies~\cite{chan2020basicvsr} have pinpointed two critical steps of VFH approaches, \emph{i.e.}, \emph{alignment} and \emph{aggregation}. 
Previous methods typically conduct pairwise alignment and aggregate frames with estimated optical flow or deformations. Due to the large pose or expression changes in face video snapshots, these approaches may fail to establish frame-wise correspondences accurately especially when input videos are in very low resolution. Furthermore, the pairwise aggregation manner neglects the long-range temporal information offered by an integral sequence but faces the risk of accumulating alignment errors.
Last but not least, to the best of our knowledge, previous VFH methods often rely heavily on beforehand aligned faces in terms of facial landmarks. Considering that raw faces in LR videos are naturally tiny and blurry~\cite{yu2019semantic}, this would drastically hinder the accuracy of face alignment. To date, how to effectively align and aggregate both spatio-temporal and facial prior information for VFH remains challenging.

In this paper, we propose a new video face hallucination method, dubbed VidFace, to super-resolve unaligned tiny face snapshots. VidFace is a full-transformer solver addressing alignment and aggregation in a unified scheme. It capitalizes on the contextual modeling ability of the attention mechanism to harness \emph{integral} information of \emph{all} the given snapshots, both spatially and temporally, for face hallucination. 
Specifically, VidFace starts with splitting snapshots into overlapped patches and flattening them into spatio-temporal tokens.
Inspired by the structural design of Token-to-Token network, we aggregate neighboring tokens iteratively to enrich the feature representation of facial details.
To further utilize the facial structural information, we propose a novel recurrent landmark positional encodings (RLPE) to equip our transformer with facial priors. RLPE is designed to recurrently estimate the landmarks and attentions together in a mutual-promotion manner, which regularises the feature alignment and aggregation without the need of pretraining transformers. 

We also present a new upsampling module, dubbed Detoken Upsampling (DeU). Apart from previous upsampling such as deconvolution and pixelshuffle that operated on each local patch, we design a novel ``Detoken Transformer'' to decode tokens into dilated ones to enlarge the latent feature resolution and then reconstruct these decoded tokens into HR images. In such dilation process, facial structure information is naturally incorporated with the help of global attention and landmark positional encoding, thus leading to superior face hallucination performance. 

Overall, our main contributions are summarized as follows:

\begin{itemize}
\vspace{-1em}
    \item We present a unified pure transformer-based solver, namely VidFace for video based face hallucination by utilizing the long-range spatial and temporal information from inputs integrally. To the best of our knowledge, we are the first attempt to super-resolve face videos by a pure transformer-based solver.
    \item We propose a novel recurrent landmark positional encoding (RLPE) to equip our transformer with facial priors. RLPE is designed to recurrently estimate the landmarks and attentions together in a mutual promotion manner, significantly facilitating facial feature alignment and aggregation.
    \item We curate a new large-scale benchmark dubbed TUFS-145K, and LR faces are unaligned in TUFS-145K, posing a more realistic and challenging scenario for existing baselines.
    \item Extensive experiments demonstrate the new state-of-the-art face hallucination performance of the proposed VidFace.
\end{itemize}

\section{Related Work}

\textbf{Video Super-resolution.} 
Dong \textit{et al.} proposed a seminar idea to tackle image super-resolution (ISR) with a Convolutional Neural Network~\cite{dong2014learning}. Based on this pioneering work, many deep learning-based approaches have been proposed for video super-resolution (VSR) task~\cite{wang2019edvr,caballero2017real,xue2019video,Tao_2017_ICCV,kappeler2016video,jo2018deep,tian2020tdan}. Comparing to ISR, VSR lays more emphasis on exploiting temporal alignment across frames. Some methods~\cite{kappeler2016video,caballero2017real,Tao_2017_ICCV} utilize optical flow to estimate the motions between frames. Xue \textit{et al.} proposed TOFlow~\cite{xue2019video}, an end-to-end trainable network to predict task-oriented motion and fuse all input frames according to estimated motion fields. Besides, several methods~\cite{jo2018deep,wang2019edvr,tian2020tdan} supplant flow estimation with implicit alignment mechanism. 
TDAN~\cite{tian2020tdan} employs deformable convolution to align neighbouring frames at the feature level. Based on~\cite{tian2020tdan}, EDVR~\cite{wang2019edvr} improves previous works with a Pyramid, Cascading and Deformable (PCD) alignment module to align multi-frames. 

\textbf{Face Hallucination.}
Face hallucination, a domain-specific super-resolution task, aims at generating high-resolution facial images from low-resolution inputs. Based on deep learning, face super-resolution methods have been actively researched and achieved impressive progress. Yu \textit{et al.} \cite{yu2016ultra} super-resolve aligned tiny face images with GAN-based models. In order to process unaligned face images, Yu \textit{et al.} \cite{yu2017face} insert multiple spatial transformer networks (STN) in a generator to hallucinate LR images. Cao \textit{et al.} \cite{cao2017attention} learn the recurrent policy network and local enhancement network through deep reinforcement learning. Zhang \textit{et al.} \cite{zhang2020copy} propose a two-branch super-resolution network (CPGAN) to upsample face images while compensating for low and nonuniform illumination. 

\textbf{Vision Transformer.} 
Transformer models have been heralded as a breakthrough in Natural Language Processing (NLP) domain, and are gradually applied to a wide range of computer vision tasks such as semantic segmentation~\cite{luo2018macro,luo2019significance,luo2021category,luo2019taking} due to its feature extraction ability and impressive performance. Yang \textit{et al.} \cite{yang2020learning} formulated LR and Ref images as queries and keys in a transformer, which contained a hard-attention module and a soft-attention module to discover deep feature correspondence and transfer rich textures to help super-resolving the input LR image. Different from Yang \textit{et al.} \cite{yang2020learning}, we formulate central frames and other frames as queries and keys in order to search relevant regions between them and fuse the correspondent feature in a pyramid transformer block.

Our work does not follow the strategy of using local-perceptual operator as optical flow or deformable convolution. Instead, a pure-transformer-based method is proposed to harness global information spatially and temporally to exploit better alignment. To our knowledge, we are the first works to propose a full-transformer solution for VFH.

\section{Methodology}

\subsection{Problem Setting and Overall Idea}\label{sec_Problem setting}

Given $N$ unaligned tiny low-resolution face snapshots captured from human video, denoted as $\{\bm{LR}_{n}\}$ where $n \in \{1, ..., N\}$, we aim at super-resolving the reference frame $\bm{LR}_{ref} \in \{\bm{LR}_{n}\}$ to restore the facial details. Our task can be clearly distinguished from traditional facial video super-resolution in two folds. First, our method operates on tiny inconsecutive data, \emph{e.g.}, each snapshot is sized $16 \times 12$ with large pose or expression divergence. Second, the faces are not aligned to their landmarks across snapshots by any pre-process. To our knowledge, such tiny, inconsecutive and unaligned facial snapshots pose a practical scenario that challenges all existing efforts. For convenience, we denote other snapshots as auxiliary snapshots $\bm{LR}_{aux}$, which are utilized to supplement the reference one $\bm{LR}_{ref}$ in hallucination. 

Our overall idea is to capitalize on the contextual modeling ability of attention mechanism to harnesses all the spatial, temporal and facial prior information of given face video snapshots. We hence propose the VidFace, an unified transformer-based solver for VFH, which boosts existing methods in all the \emph{alignment}, \emph{aggregation} and \emph{upsampling} steps. We detail the VidFace in what follows.

\subsection{Network Architecture}

\begin{figure}
  \centering
  \includegraphics[scale=0.43]{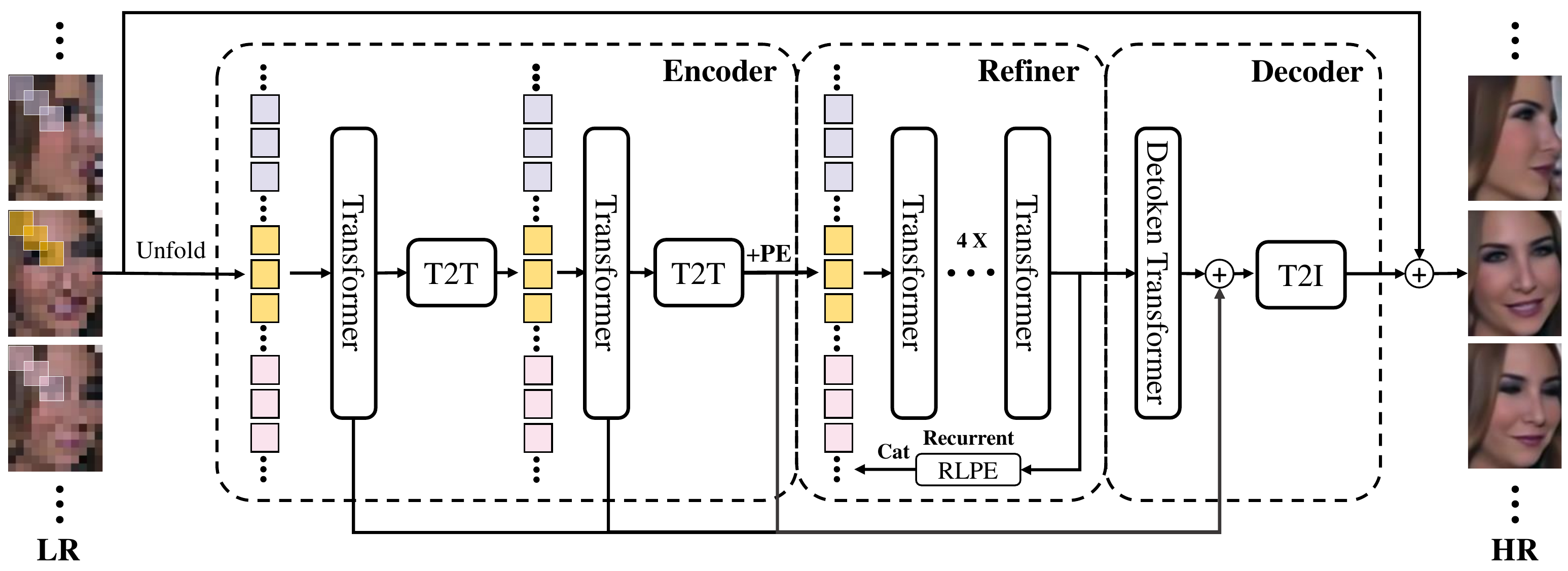}
  \caption{The network architecture of VidFace.}
  \label{fig_VidFace}
\end{figure}

The network architecture of VidFace is illustrated in Fig.~\ref{fig_VidFace}. It is composed of an encoder $\mathbf{E}$, a refiner $\mathbf{R}$ and a decoder $\mathbf{D}$, wherein $\mathbf{E}$ is responsible for an initiatory spatio-temporal feature alignment and aggregation, $\mathbf{R}$ recurrently refines the latent features with facial structure prior and $\mathbf{D}$ generates the final HR face with a facial structure-aware upsampling. 

\textbf{Encoder as Token-to-token Transformer.} 
The design spirit of the encoder $\mathbf{E}$ is inspired by Token-to-Token (T2T) ViT but also introduces many improvements tailored for VFH. Given $N$ face video snapshots $\{\bm{LR}_{n}\}$ where $n \in \{1, ..., N\}$, we first $\texttt{Unfold}$ them to softly split snapshots into overlapped patches. These patches are then flattened in to tokens $\bm{T} = \{\bm{T}_n\}$ (where $\bm{T}_n$ denotes the tokens in $\bm{LR}_{n}$) and forwarded to transformer layer for feature extraction. Finally, the output tokens from transformer are restructured as an image again to bootstrap the next unfold-reconstruction iteration. For a (restructured) LR snapshot $\bm{LR}_n\in \mathbb{R}^{h\times w\times c}$, the number of output tokens can be calculated as:
\begin{eqnarray}
l_n=\left \lfloor \frac{h+2p-k}{s} + 1\right \rfloor \times \left \lfloor \frac{w+2p-k}{s} + 1\right \rfloor, 
\end{eqnarray}
where $k$ is the size of patch, $s$ is the stride and $p$ is the padding on $\bm{LR}_n$. Then the total number of tokens in $N$ snapshots will be $L = \sum_{n=1}^{N} l_{n}$.
Different from vanilla T2T process which reduces the number of tokens progressively, VidFace keeps same amount of tokens to maintain the resolution during encoding. VidFace also stacks the tokens in different layers to further enrich the feature with multi-scale representation. As the neighboring (either intra- or inter-snapshot) tokens are aggregated iteratively into single token, the spatio-temporal information are naturally embedded into them. Moreover, the interaction between overlapped tokens also helps to extract more structure information like edges and textures, which is of pivotal for facial detail hallucination.

In order to maintain the spatial structure of snapshots when splitting tokens, we add the spatial position encoding to tokens as in vanilla transformer. It is noteworthy that we do not consider temporal position encoding since we assume that the ``snapshots'' are unordered compared to ``frames''.

\textbf{Refiner as Recurrent Landmark Transformer.} 
We further refine the initiatory spatio-temporal tokens by introducing facial structure information, which has been proven as a robust prior to regularise the feature alignment \& aggregation~\cite{yu2017face,zhang2018super,grm2019face}. However, to utilize such prior effectively in a transformer is non-trivial. Here we propose to additionally estimate the facial landmarks during the training and employ the estimated facial landmarks in positional encoding. To achieve the auxiliary landmark estimation task, we first utilize a linear projection to decrease the dimension of every tokens. Then all tokens from the same snapshot will be flattened into a one-dimensional feature. Followed by a linear layer and a landmark head, these $5$ landmark positions of the face will be predicted as:

\begin{eqnarray}
\bm{LM}_{n} = \texttt{LMHead}(\texttt{MLP}(\texttt{Flatten}(\bm{T}_{n,0},...,\bm{T}_{n,l_n}))),
\end{eqnarray}
where $l_n=h\times w$ denotes the tokens number of snapshot $LR_n$.

To represent the positions encoding of each token in terms of landmarks, we employ the estimated facial landmarks as the anchors. The landmark position encoding of token $\bm{T}_{n,i}$ is denoted by $\bm{RLPE}_{n, i} \in \mathbb{R}^{5}$, which is reflected in the distance between token $\bm{T}_{n,i}$ and $5$ facial landmarks. Concretely, $\bm{RLPE}_{n, i}$ can be calculated as:

\begin{eqnarray}
\bm{RLPE}_{n, i} = [d(\bm{P}_{n, i}, lm_{n, 1}), ..., d(\bm{P}_{n, i}, lm_{n, 5})],
\end{eqnarray}
where $d(.)$ indicates the \emph{Euclidean} distance between two points in a snapshot, $\bm{P}_{n, i}$ is the position of token $\bm{T}_{n, i}$ and $\bm{lm}_{n, j}$ indicates the \emph{j}-th facial landmark of the \emph{n}-th snapshot obtained by the recurrent transformer, which serves as the anchor point when calculating $\bm{RLPE}_{n,i}$. In order to handle faces that occupy different sizes in images, we normalize $\bm{RLPE}$ (\emph{i.e.}, $\frac{\bm{RLPE}}{\left \| \bm{RLPE} \right \|_{1}}$ ) before concatenating it to the tokens. Since the auxiliary head is trained from scratch and the estimated landmarks are inaccurate in early iteration, we propose to recurrently learn the facial landmark and attention map together in a mutual-promotion manner. Specially, the yielded landmarks would recurrently embedded into the position encoding, together with the refined tokens to bootstrap the next refine iteration, as shown in Fig.~\ref{fig_VidFace}. Afterwards, the final tokens would be forwarded to the decoder.

\textbf{Decoder as Detoken Upsampling.}
We propose Detoken Upsampling (DeU) to conduct the facial structure-aware upsampling, which reconstructs the latent tokens into $8\times$ facial images, expressed as:

\begin{eqnarray}
\bm{T}^{l\times d_{model}} \mapsto \bm{HR}^{8h \times 8w \times 3},
\end{eqnarray}
where $d_{model}$ represent the input channels, respectively. DeU is composed of two key modules, \emph{i.e.}, \emph{Detoken Transformer (DeT)} and \emph{Tokens to Image (T2I)}.

\emph{Detoken Transformer (DeT)} is designed to overcome the limitation of deconvolution~\cite{noh2015learning} or sub-pixel convolution~\cite{shi2016real} on their inherent local perceptivity. 
DeT decodes tokens into dilated ones to enlarge the latent feature resolution. The feature of every token has the dimension of $d_{out}*k*k$, where $k$ is the kernel size used in T2I and $d_{out}$ is the output dimension of T2I. In such dilation process, facial structure information is naturally employed with the help of global attention and landmark positional encoding.

\begin{wrapfigure}{r}{8.5cm}
\centering
\includegraphics[width=0.5\textwidth]{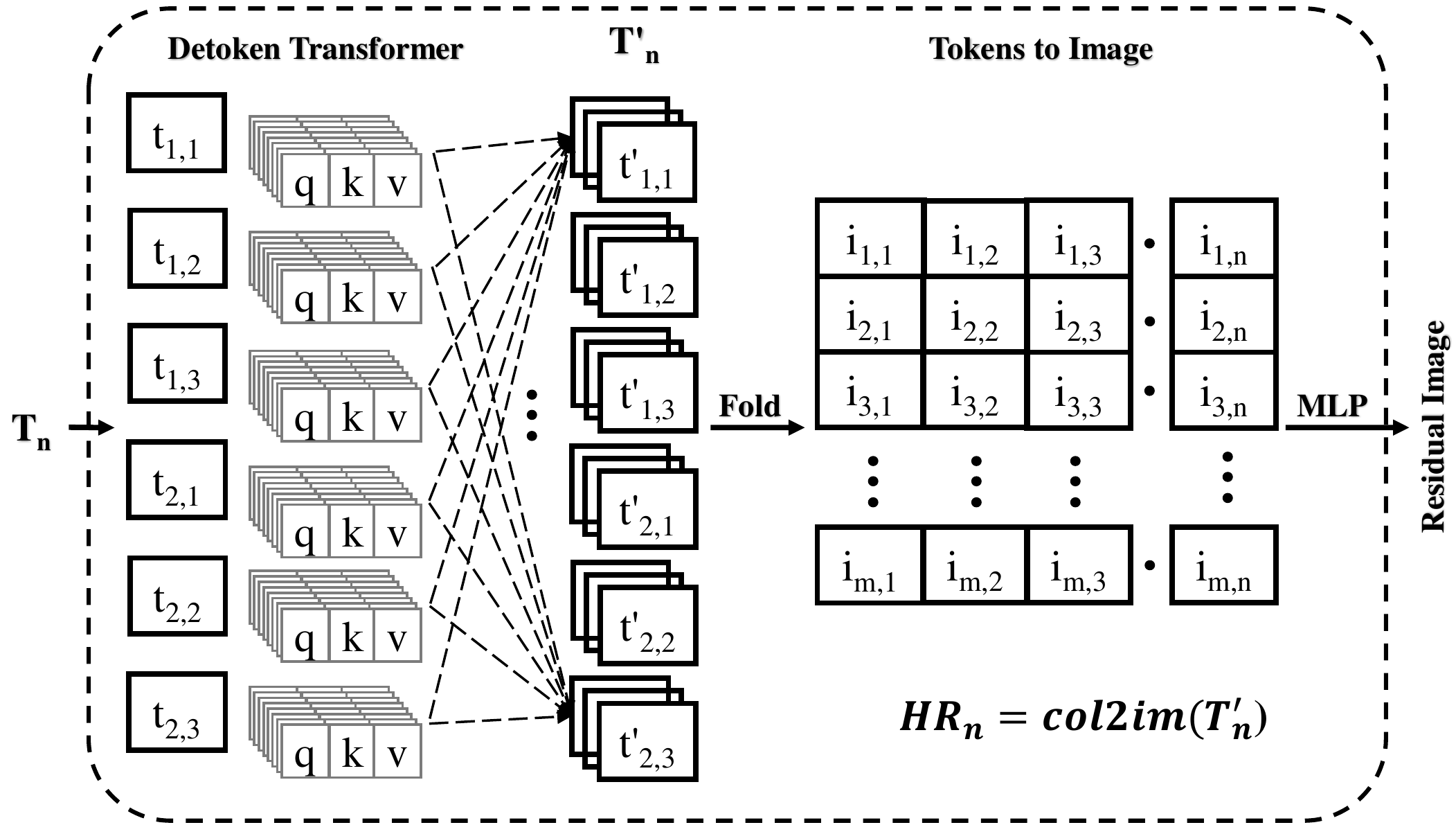}
\caption{\footnotesize The architectural overview of DeU.}
\label{fig_DeU}
\end{wrapfigure}

As depicted in Fig.~\ref{fig_DeU}, for video tokens $\bm{T}$ whose feature dimension is $d_{model}$, we utilize multi-head attention~\cite{vaswani2017attention} to extract the $d_{q}$-dimensional queries, $d_{k}$-dimensional keys and $d_{v}$-dimensional values for $h$ times with different, learnable linear projections. We remove the vanilla feed forward layer after the multi-head attention to maintain the global perceptual field. Formally, the self-attention process can be presented in the following function:

\begin{equation}
\begin{aligned}
DeT(\bm{T}) &= \texttt{Concat}(\texttt{head}_{1}, \dots,\texttt{head}_{h}) + \texttt{Concat}(\bm{TW}_{1}^{V},\dots,\bm{TW}_{h}^{V}),\\
where \ \texttt{head}_{i} &= \texttt{Attention}(\bm{TW}_{i}^{Q}, \bm{TW}_{i}^{K}, \bm{TW}_{i}^{V}),
\end{aligned}
\end{equation}

where the projection matrix $\bm{W}_{i}^{Q}\in \mathbb{R}^{d_{model}\times d_{q}}$, $\bm{W}_{i}^{K}\in \mathbb{R}^{d_{model}\times d_{k}}$, $\bm{W}_{i}^{V}\in \mathbb{R}^{d_{model}\times d_{v}}$. For instance, when using $d_{q} = d_{k} = 32$, $d_{v}=1024$ and $h = 8$, the token feature would be extended, thus can be then folded to a $8\times$ feature map using T2I.

\emph{Tokens to Image (T2I)} refers to the eventual step for face hallucination to reconstruct tokens into images. We take the tokens $\bm{T}'_{n} \in \bm{T}'$ from snapshot $n$ as an example to illustrate T2I as shown in Fig.~\ref{fig_DeU}, where $\bm{T}'$ denotes the overall video tokens generated from DeT. The reconstruction process can be formatted as:
\begin{eqnarray}
\bm{HR}_{n} = \texttt{Reshape}(\texttt{MLP}(\texttt{T2I}(\bm{T}_{n}'))).
\end{eqnarray}

Inspired by the implementation of deconvolutional layer, we implement \texttt{T2I} with \texttt{fold} (a.k.a, \emph{col2im}) as follows:

\begin{eqnarray}
\texttt{T2I}(\bm{T}_{n}') = \texttt{Fold}(\bm{T}_{n}', k, s, p),
\end{eqnarray}
where $k, s, p$ denote the kernel size, stride and padding used in \texttt{Fold}, respectively. For instance, when setting $k=16$, $s=8$ and $p=4$ as in our work, we can upsample LR snapshots into $8\times$ HR faces which can be regard as an inverse operation of T2T.

\subsection{Training Objective}
The training objective of VidFace is twofold. For the main task of face hallucination, we adopt the Charbonnier Loss \cite{li2020mucan} to produce better refined edges during training. Specially, Charbonnier Loss is defined as:

\begin{equation}
\begin{aligned}
\mathcal{L}_{pix} = \sqrt{{\left \| \hat{\bm{HR}}_{n} - \bm{HR}_{n}\right \|} ^{2} + \epsilon^{2}} \,,
\end{aligned}
\end{equation}
where $\hat{\bm{HR}}_{n}$ is the predicted high-resolution result, and $\epsilon$ is a small constant. 

For the auxiliary task of landmark estimation, we use a robust loss function (smooth-L$_1$) defined in~\cite{girshick2015fast} to optimize the five facial landmarks. Specially, smooth-L$_1$ can be defined as $L_{pts} (lm_n, lm^{*}_n)$, where $\bm{lm}_n=\{lm_{x_1}, lm_{y_1}, \dots , lm_{x_5}, lm_{y_5}\}_n$ and $\bm{lm}^{*}_n=\{lm^{*}_{x_1}, lm^{*}_{y_1}, \dots , lm^{*}_{x_5}, l^{*}_{y_5}\}_n$ represent the predicted five facial landmarks and ground-truth. 

\begin{equation}
\mathcal{L}_{pts} (\bm{lm}_n, \bm{lm}^{*}_n)=\left\{\begin{matrix} 0.5 \left ( \bm{lm}_{n}-\bm{lm}_{n}^{*} \right )^{2},
 & if\left | \bm{lm}_{n}-\bm{lm}_{n}^{*} \right | <1 \\ \left | \bm{lm}_{n}-\bm{lm}_{n}^{*} \right | -0.5,
 & otherwise
\end{matrix}\right.
\end{equation}
Therefore, the final loss is formulated as:
\begin{equation}
\begin{aligned}
\mathcal{\bm{L}}_{total} = \sum_{n}\left (\lambda_{1} \mathcal{L}_{pix}+\lambda_{2} \mathcal{L}_{pts} \right),
\end{aligned}
\end{equation}
where $\lambda_{1}$ and $\lambda_{2}$ denote hyper-parameters controlling the relative importance of the two losses.

\section{Experiments}
\label{sec_exp}

To demonstrate the advantages of VidFace, we have conducted extensive experiments on both public and our new curated datasets. We first briefly introduce the evaluation datasets, the corresponding evaluation protocols, and the implementation details. Then, we perform comprehensive comparisons to verify the superiority of VidFace over all the compared state-of-the-art approaches. Finally, we perform in-depth analysis on each module with ablation study to pinpoint their contributions for VFH.

\subsection{Datasets and Evaluation Protocols}

\begin{figure}
\centering
\includegraphics[width=0.99\textwidth]{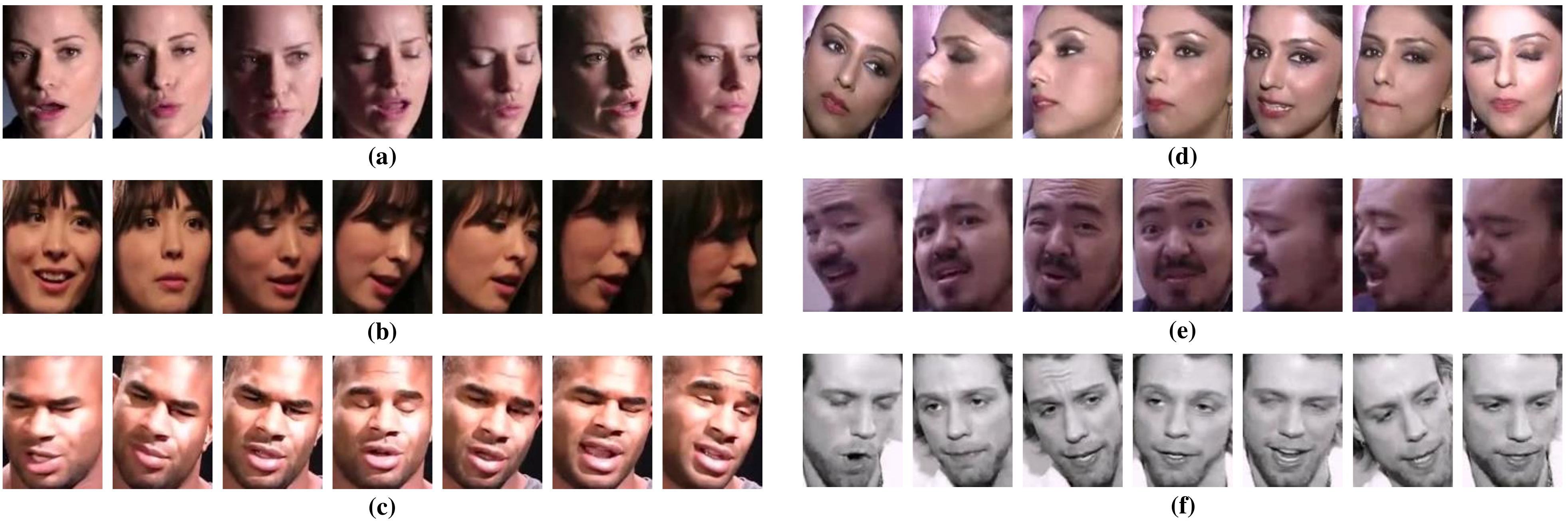}
\caption{\footnotesize Some facial video snapshot examples of TUFS-145K.}
\label{fig_tufs}
\end{figure}

We employ two datasets for a comprehensive evaluation to validate the effectiveness of VidFace. \textbf{IJB-C} is a video-based face database which contains $3,531$ subjects with $31.3K$ still images and $117.5K$ frames from $11,779$ videos. We set a threshold to filter out the over-short video whose length is less than $4$ and results in a subset with $6.4K$ video clips. \textbf{TUFS-145K} is origin from the public \emph{voxceleb2}~\cite{chung2018voxceleb2} dataset. We first detect and crop the face boundingbox of the videos every several frames with pyramidbox tool. We score the candidate clips with (1) faceQnet to judge the quality of cropped faces and (2) 3ddfa to estimate the pose variance. By combined considering  these two scores, we select 145K 7-frame clips from the candidates to constitute TUFS-145K, which has favourable face quality and large pose variance. TUFS-145K is divided into a training set of $116K$ video clips, a validation set of $14.5K$ clips  and a testing set of $14.5K$ clips. During the network training, we employ Bicubic downsampling to generate the LR face video snapshots as the input data. 

The estimated high-resolution images are evaluated by peak signal to noise ratio (PSNR) and the structural similarity index (SSIM) on $Y$ channel (\emph{i.e.}, luminance) of the transformed YCbCr space and RGB space.

\subsection{Implementation Details}
We use PyTorch~\cite{paszke2017automatic} for our implementation. We employ T2T-ViT~\cite{yuan2021tokens} as the backbone. In order to stablize the training, we choose  AdamW~\cite{loshchilov2018fixing} and Lookahead~\cite{zhang2019lookahead} as the optimizer. In addition, we set 0.05 as the maximum $l_{2}$ norm for gradient clipping. We train the network for a total of $600k$ iterations. And the learning rate is decreased by cosine annealing scheduler~\cite{loshchilov2016sgdr} without restart. We resize the snapshots to $16 \times 12$ with Bicubic operation as the LR input and the length of the snapshots is set to $7$. These frames are shuffled in training to improve the generalization. The batchsize is set to $16$, which is the same with MUCAN~\cite{li2020mucan} and EDVR\cite{wang2019edvr} in all experiments for fair comparison. 

\begin{table}[h]
\begin{center}
\caption{Quantitative evaluation on TUFS-145K and IJBC dataset.}
\label{tab_main}
\vspace{-2mm}
\scalebox{0.9}{
\begin{tabular}{lcccccccccccc}
\toprule
\multirow{4}{*}{\vspace{-2mm}Model} &
\multicolumn{5}{c}{Y-channel} & & \multicolumn{5}{c}{RGB} & \\
\cmidrule{2-6} \cmidrule{8-12}
& \multicolumn{2}{c}{TUFS-145K} & & \multicolumn{2}{c}{IJB-C} & & \multicolumn{2}{c}{TUFS-145K} & & \multicolumn{2}{c}{IJB-C} \\
\cmidrule{2-3} \cmidrule{5-6} \cmidrule{8-9} \cmidrule{11-12} 
& PSNR & SSIM & & PSNR & SSIM & & PSNR & SSIM & & PSNR & SSIM \\ 
\hline
Bicubic  & 25.72 & 0.7914 &  & 23.86 & 0.7029 & & 24.31 & 0.7631 & & 22.44 & 0.6647 \\
EDVR~\cite{wang2019edvr}  & 30.55 & 0.9119 &  & 26.05 & 0.7968 & & 29.08 & 0.8957 & & 24.59 & 0.7675\\
MUCAN~\cite{li2020mucan} & 30.49 & 0.9113 &  & 26.11 & 0.7988 & & 29.00 & 0.8949 & & 24.64 & 0.7693\\
VidFace (Ours)  & \textbf{30.94} &\textbf{ 0.9181} &  & \textbf{26.35} & \textbf{0.8076} & & \textbf{29.46} & \textbf{0.9026} & & \textbf{24.89} & \textbf{0.7795}\\

\bottomrule
\end{tabular}
}
\end{center}
\end{table}

\begin{figure}[t]
\centering
\includegraphics[width=0.99\textwidth]{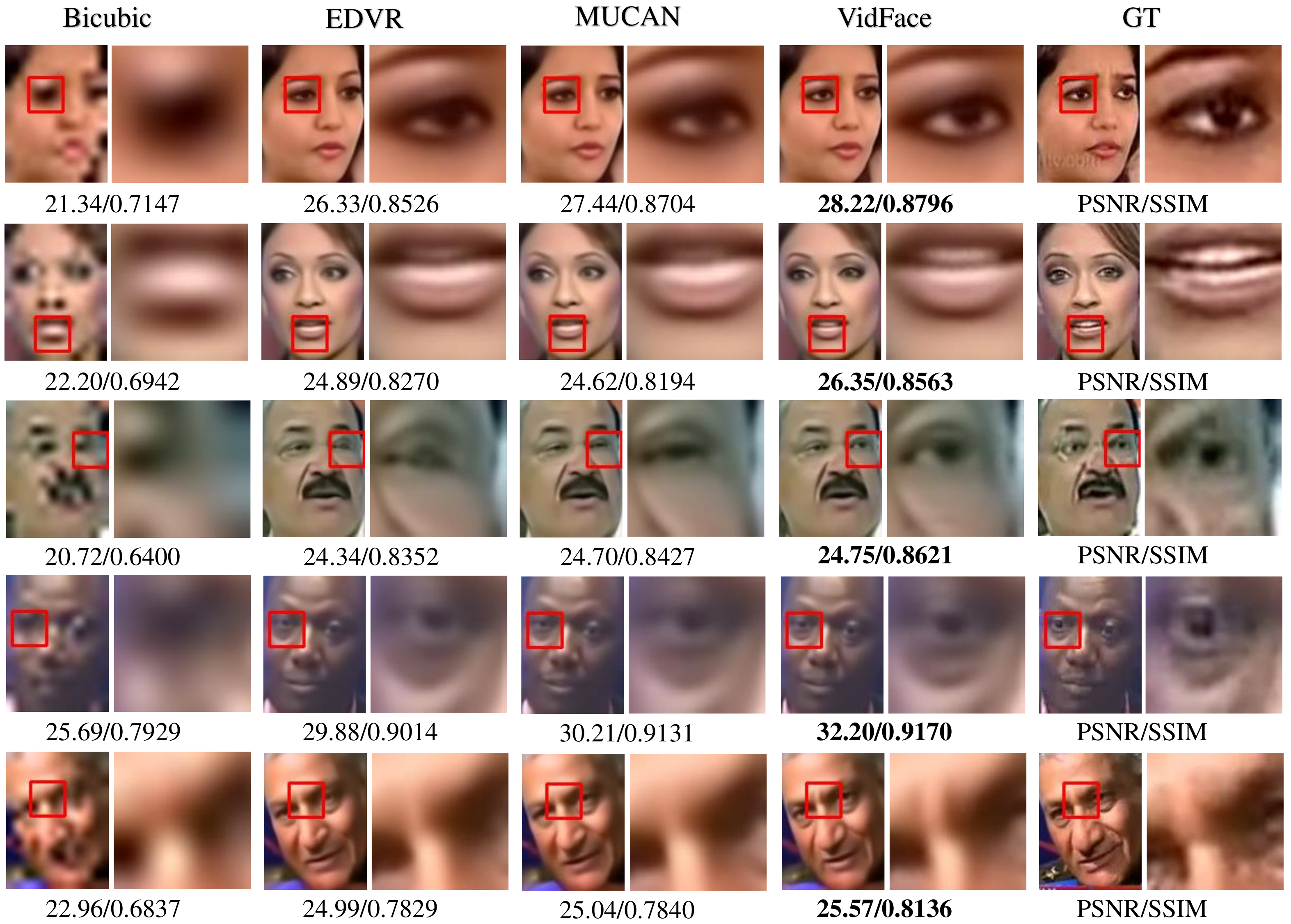}
\caption{\footnotesize Qualitative comparison on the TUFS-145K dataset for $8\times$ video face snapshot SR. The proposed VidFace
hallucinates more facial details specially in ``eyes'', ``mouth'' and so on. Zoom in for best view.}
\label{fig_res1}
\end{figure}

\subsection{Comparison with State-of-the-Art Methods}
We compare our method with several state-of-the-art methods, including EDVR~\cite{wang2019edvr} and MUCAN~\cite{li2020mucan}. EDVR is a typical deformable convolution-based method. MUCAN depends on the self-attention modules for alignment and aggregation, but limited to pairwise frames. It is noteworthy that we have not reported the results of optical flow-based methods since these methods are not applicable for over-small unaligned snapshots (the spatial pyramid used in optical flow calculation requires the image to be larger than $16 \times 16$). The hallucination results of Bicubic are also reported. The PSNR and SSIM comparisons between VidFace and the above mentioned methods on TUFS-145K as well as IJB-C by $8\times$ upscaling are shown in Table~\ref{tab_main}. The average Y-channel and RGB-space PSNR and SSIM on all the evaluation snapshots are reported. 

Firstly, our proposed VidFace achieves the highest PSNR, yielding $30.94 dB$ on TUFS-145K and $26.35 dB$ on IJB-C and significantly outperforming EDVR over $0.4 dB$ and $0.2 dB$, which demonstrates the superiority of a full-transformer base solver over the conventional deformable convolutional. The similar result can be also observed on SSIM score. 

\begin{figure}[t]
\centering
\includegraphics[width=0.99\textwidth]{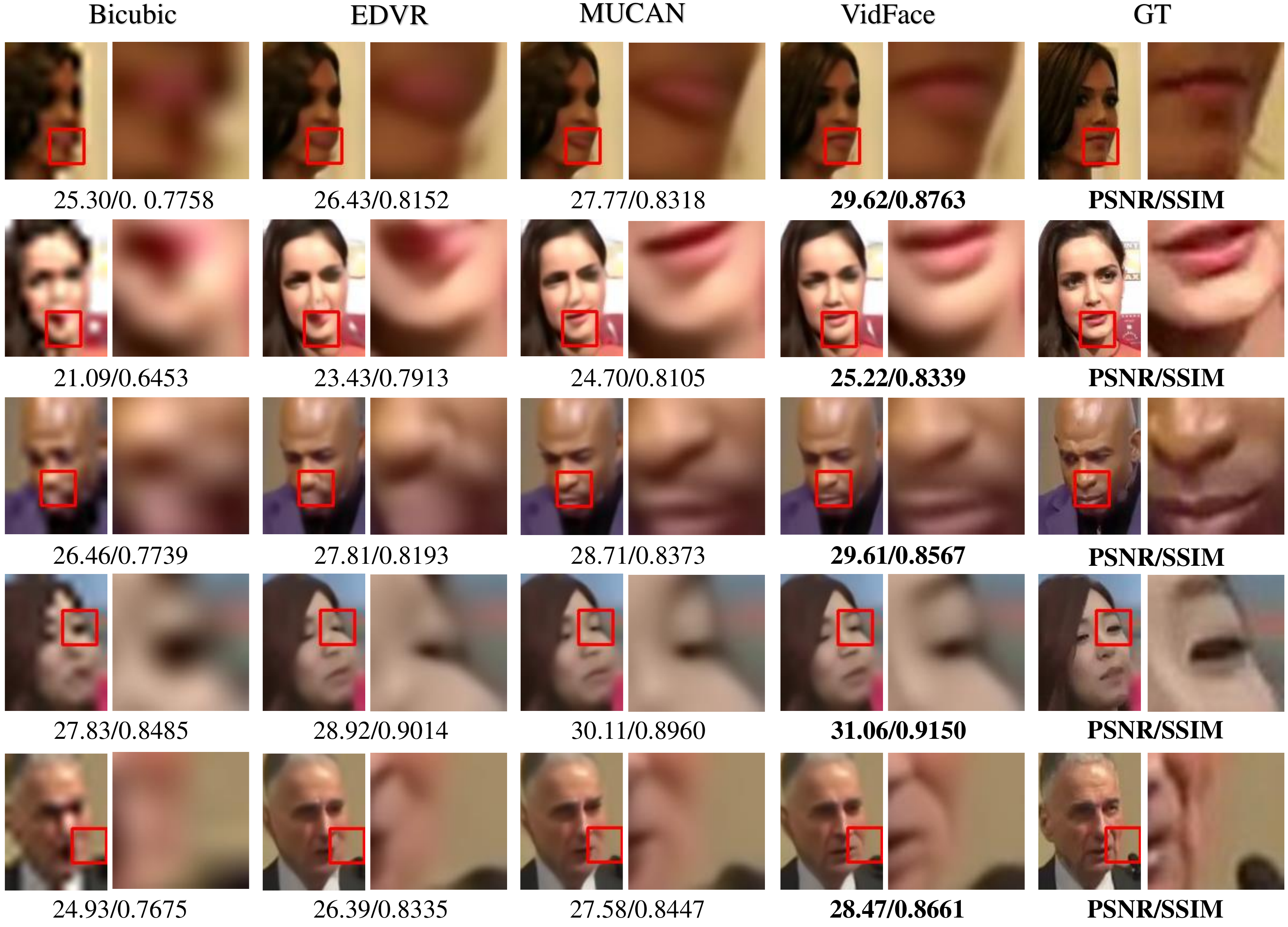}
\caption{\footnotesize Qualitative comparison on the IJB-C dataset for $8\times$ video face snapshot SR. The proposed VidFace hallucinates more facial details specially in ``eyes'', ``mouth'' and so on. Zoom in for best view.}
\label{fig_res2}
\end{figure}

Secondly, VidFace significantly outperforms MUCAN by $0.45 dB$ and $0.24 dB$. Since the alignment mechanism of MUCAN also relies on spatial self-attention, such better results draw our attention to the temporal space. Comparing to MUCAN, VidFace handles multiple snapshots all at once and harnesses the spatial and temporal information integrally, thus boost the alignment and aggregation. We also give the visualization results on the above two datasets shown in Fig.~\ref{fig_res1} and Fig.~\ref{fig_res2}. These qualitative result further verifies that VidFace can actually boost the final face hallucination.

\begin{figure}[t]
\centering
\includegraphics[width=0.99\textwidth]{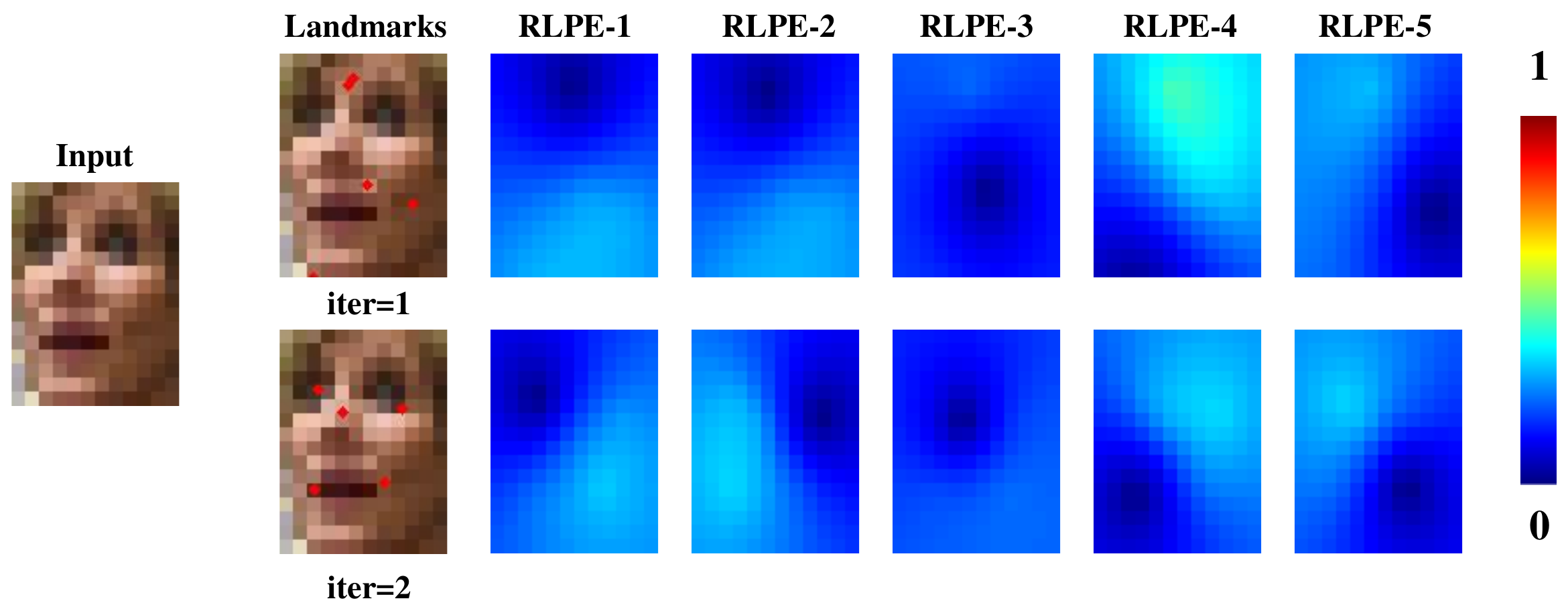}
\caption{\footnotesize The visualization of RLPE. We show both the estimated landmarks and the position encoding in terms of distance field. The landmarks are estimated from scratch, which would be increasing accurate after iterations thus recurrently yielding finer facial prior information to regularize the position encoding for transformer.}
\label{fig_rl}
\end{figure}

\subsection{Ablation Study}
To assess the importance of various aspects of the VidFace, we run experiments on TUFS-145K under the setting of Deformable convolution-based encoder, T2T transformer-based encoder, RLPE refiner and DeU decoder, deactivating one or a few modules at a time while keeping the others activated. Table~\ref{tab_ablation} reports the face hallucination performance in terms of PSNR under different ablations. To begin with, we compared the deformable and transformer-based backbones. As observed in the table, the employment of global spatio-temporal information in transformer brings significant improvement ($0.15 \uparrow$) to PSNR score. Besides, the employment of RLPE and DeU brings about another $0.23 \uparrow$ gain in PSNR. Together with the visualization result of RLPE depicted in Fig.~\ref{fig_rl}, we can conclude that the designed modules are effective for face hallucination.
More experimental results can be found in the Appendix.

\begin{table}[t]
    \caption{\textbf{Ablation studies of the components.} Each component brings significant improvements in PSNR, verifying their effectiveness.}
    \label{tab_ablation}
    \begin{center}
        \tabcolsep=0.13cm
        \scalebox{0.85}{
            \begin{tabular}{l|c|c|c}
                \hline
                                           & (A)            & (B)         & \textbf{VidFace}       \\\hline
                Deformable Conv            &  \checkmark    &             &                        \\
                T2T Transformer            &                &  \checkmark & \checkmark             \\
                DeU                        &                &             & \checkmark             \\
                RLPE                       &                &             & \checkmark             \\\hline\hline
                PSNR (dB)                  & 29.08          & 29.23 (0.15$\uparrow$)      &  29.46 (0.38$\uparrow$)                  \\\hline
            \end{tabular}}
    \end{center}
\end{table}

\section{Conclusion}
This work devotes attention to jointly model both spatio-temporal and facial prior information for better video face hallucination with unaligned tiny snapshots. We pinpoint and analyze the shortcomings of previous work and accordingly propose the VidFace. VidFace is a pure-transformer method to address alignment, aggregation and upsampling in an unified scheme. We further curate a large-scale benchmark of tiny unaligned facial video snapshots, dubbed TUFS-145K, which poses more realistic and challenging scenario for existing baselines and hopefully advances the future face hallucination research. To our knowledge, we are the first attempt to propose a unified transformer-based solver tailored for VFH task. Extensive experiments on video face benchmarks show that the proposed method significantly outperforms the state of the arts.

\end{document}